\def\year{2015}
\documentclass[letterpaper]{article}
\usepackage{aaai}
\usepackage{times}
\usepackage{helvet}
\usepackage{courier}
\usepackage{amsmath,amssymb,amsthm}
\usepackage{color}
\usepackage{graphicx}

\theoremstyle{definition} 
\newtheorem{definition}{Definition} 
\theoremstyle{remark}
\newtheorem*{remark}{Remark}

\newcommand{\mc}[1]{\mathcal #1}
\newcommand\harp[1]{\mathstrut\mkern2.5mu#1\mkern-11mu\raise0.6ex%
  \hbox{$\scriptscriptstyle\rightharpoonup$}}
\newcommand\reducesto{_{\harp{r}}}
  
\newcommand\independent{\protect\mathpalette{\protect\independenT}{\perp}}
\def\independenT#1#2{\mathrel{\rlap{$#1#2$}\mkern2mu{#1#2}}}
\newcommand\notindependent{\not\!\perp\!\!\!\perp}

\begin{document}
%
\title{Using Bayesian Network Representations \\for Effective Sampling from Generative Network Models}
\author{Pablo Robles-Granda \and Sebastian Moreno \and Jennifer Neville\\
Computer Science Department\\
Purdue University\\
West Lafayette, IN 47907\\
}

\maketitle
\begin{abstract}
\begin{quote}
Bayesian networks (BNs) are used for inference and sampling by exploiting conditional independence among random variables.
Context specific independence (CSI) is a property of graphical models where additional independence relations arise in the {\em context} of particular values of random variables (RVs). Identifying and exploiting CSI properties can simplify inference. 
Some generative network models (models that generate social/information network samples from a network distribution $P(G)$), with complex interactions among a set of RVs,
 can be represented with probabilistic graphical models, in particular with BNs. 
In the present work we show one such a case. We discuss how a mixed Kronecker Product Graph Model can be represented as a BN, and study its BN properties that can be used for efficient sampling. Specifically, we show that instead of exhibiting CSI properties, the model has {\em deterministic context-specific dependence} (DCSD). Exploiting this property focuses the sampling method on a subset of the sampling space that improves efficiency.

\end{quote}
\end{abstract}

\section{Introduction}

In the last few decades Bayesian networks (BNs)~\cite{pearl:88} have grown from a theoretical approach to model joint distributions, to a powerful tool that can be applied to solve many real-world problems due to the relative ease of estimation and inference. Specifically, a BN is a directed acyclic graph where nodes represent random variables (RVs) and edges represent conditional dependence of variables in the direction specified in the graph.

One of the most important characteristics of BNs is the \emph{relative} ease of the inference process. For instance, the use of a specific context $C=c$ over a set of variables (i.e. values assigned to them) can facilitate computation of the posterior probability of the remaining variables given the context ($P(X|C=c)$) \cite{Boutilier:96}. Even though it has been demonstrated that the exact inference problem is NP-hard for arbitrary BNs~\cite{Cooper:90}, in some cases, the contextual structure can be used for tractable inference.

In addition to inference, BNs can be utilized for sampling. The sampling process generally involves determining a topological ordering of the variables (i.e., $X_1,\ldots,X_n$), then iteratively drawing the value for each RV given the previous sampled values (i.e., the context $C=c$). To draw the value of a specific RV, the methods compute the corresponding probability distribution $P(X_i|C=c)$, sample the value of the variable, add the sampled value of $x_i$ to $C$, repeating the same process up to the last variable $X_n$.

Considering the relevance of BNs and their sampling process, BNs can also be utilized to model the formation and structure of relational networks---i.e., social, information, biological networks, where nodes correspond to entities and links represent relations among the entities, (such as friendship links in Facebook). In this paper, we show that probabilistic \emph{generative network models} (GNMs)\footnote{GNMs should not be confused with probabilistic graphical models, such as Bayesian networks. To avoid confusion we will refer to probabilistic graphical models as ``graphs'', and to networks sampled from GNM as ``networks'', except for Bayesian networks which are widely known as such.} can be reduced to BNs, and BN sampling methods can be applied to generate networks. Some well known GNMs are: Erd\"{o}s-R\'{e}nyi~\cite{Erdos:59}, Chung Lu~\cite{Chung:02}, and the Kronecker product graph model (KPGM)~\cite{leskovec:10}. 

GNMs model the distribution of networks $G\!=\!(\mathbf V, \mathbf E)$ with set of nodes $\mathbf V$ and edges $\mathbf E$, through binary random variables (typically one per each possible edge in the network).  Particularly, the random variable $E_{ij}$ models the existence of an edge $e_{ij}$ between nodes $V_i\in\mathbf V$ and $V_j\in\mathbf V$, where $P(E_{ij})=\pi_{ij}$. This results in a total of $|\mathbf V|^2$ RVs. The naive sampling process of a network from a GNM samples each possible edge independently using a Bernoulli distribution. When the sample is a success (i.e., $E_{ij}=1$), then the edge $e_{ij}$ is added to the set of edges $\mathbf{E}$. Unfortunately, a naive sampling process has complexity time $O(|\mathbf V|^2)$ which make it impractical to model large networks. While there are some sampling algorithms with time complexity proportional to the number of edges ($O(|\mathbf E|)$), most of these algorithms are provably incorrect (i.e. they generate improbable networks from the underlying distribution \cite{Moreno:14}). 

Furthermore, some GNMs generate networks with properties that differ from those observed in real-world networks (e.g., transitivity, assortativity). Generating realistic random networks is important for prediction, hypothesis testing, generation of data for evaluation, randomization of sensitive data, etc. This is the motivation behind several new GNMs with more complex dependencies between the edge RVs (e.g., mKPGM \cite{Moreno:10} and BTER \cite{Seshadhri:12}).

For simple GNMs with independent binary RVs $E_{ij}$ transformation to a BN representation is not necessary. However, for some of the more recent GNMs with complex structure due to latent variables and dependencies of the edges, a BN representation can be useful to consider for sampling and inference.
Specifically, we can take advantage of existing concepts and algorithms from research on BNs, particularly from inference and learning. For example, we could (1) compactly represent the edge dependencies in the network, and (2) develop more efficient sampling mechanisms based on the conditional independence/dependence relationships encoded in the graphical model structure. 

In this paper, we consider mixed Kronecker Product Graph Models (mKPGMs) \cite{Moreno:10}. 
We show how an mKPGM can be represented as a Bayesian network with a hierarchy of latent variables that represent activations of clusters of edges at different levels in the network. Then, we consider the use of context specific independence (CSI) to facilitate the inference process and posterior sampling; however, it cannot be used to significantly reduce the time complexity of the sampling process. Then, we formalize the notion of \emph{context-specific dependence} (CSD) and \emph{deterministic context-specific dependence} (DCSD) for hierarchical GNMs. Specifically, CSD is simply CSI's complementary concept and DCSD is an extreme form (i.e., deterministic CSD). We discuss how to improve the sampling process of a GNM by exploiting the DCSD property and \emph{iteratively} sampling a hierarchy of latent variables that represent cluster activations at different levels. 

\begin{figure}[]
\begin{center}
\hspace{-5.5mm}
\includegraphics[width=1.2in]{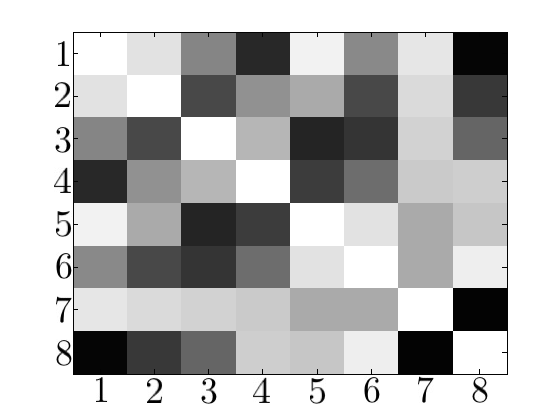}
\hspace{-4.5mm}
\includegraphics[width=1.2in]{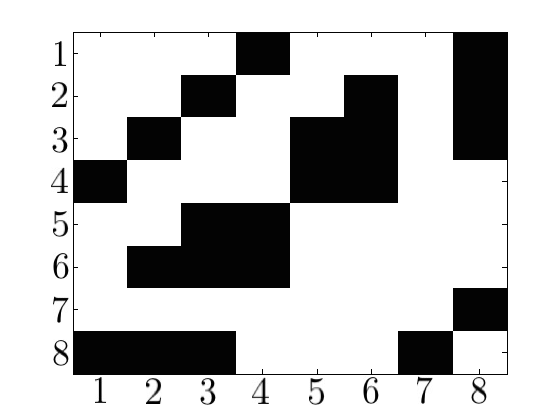}
\hspace{-4mm}
~~\includegraphics[width=1.2in]{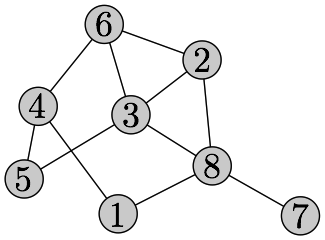}
\caption{Left: Matrix of Probabilities (grayscale depicting probability values from 0 (white) to 1 (black)). Center: Sampled adjacency matrix ($E_{ij}=0$ (white) and $E_{ij}=1$ (black)). Right: Sampled network.}
\label{fig:gnm}
\end{center}
\vspace{-4mm}
\end{figure}

\section{Background and Related Work}

Our work is related to CSI in \emph{probabilistic relational models} where the RVs are predefined. However, in our analysis we encounter a varying number of RVs and configurations as opposed to the case of probabilistic relational models. The most representative work in CSI for probabilistic relational models is that of \cite{Fierens:10}. 
Also close to our analysis is the work of \cite{Nyman:14} and \cite{Pensar:14} that deal with directed acyclic graphs and decomposable stratified graphical models, respectively.  Both works allow to reduce the size of the CPD to calculate the joint distribution. Our work does not require to calculate the joint but rather samples networks using randomization (that can be achieved through group probability sampling). 

\subsection{Bayesian Networks}

A Bayesian network BN is a directed acyclic graph where the nodes represent RVs and the edges represent (directed) dependencies between variables. More precisely,
a node in a BN is an RV that is conditionally dependent on its parents. Thus, each node in the BN has a conditional probability associated explicitly, by design. Let $X_1,X_2,\ldots,X_n$ be a topological ordering of the nodes in the BN. Then, $X_i$ is independent of $(X_1\ldots X_{i-1} \backslash pa(X_i))|pa(X_i)$. 
In consequence, the BN implicitly represents conditional independence relations. This simplifies the computation of the joint distribution of the RVs which can simply be stated as:
$$
P(X_1,X_2,\ldots,X_n)=\prod_{i=1}^{n} P(X_i|pa(X_i))
$$

\subsection{Bayesian Network Independence Properties}

The two main properties of BNs that are exploited for inference are: conditional independence (CI) and context-specific independence (CSI) \cite{Boutilier:96}. 
We describe CI and CSI (later we derive related properties CSD and DCSD), without describing the details of how particular inference algorithms use these properties for inference, to simplify the exposition. 
CI appears as the main characteristic in the structure of BNs whereby the joint distribution can be represented by focusing in the conditional dependencies of RVs.
The idea behind it is that the joint distribution can be computed more efficiently by considering the conditional independence relations of RVs which do not impact the computation and use only the relevant nodes than considering all the nodes. This leads to a more efficient estimation of the conditional probability distributions of the RVs. 
The posterior distribution of some RVs can be computed in a tractable manner when other variables are observed, because only certain variables have impact in the distribution of a node in the BN (the node's parents, its children, and its children's other parents). These variables (affecting the distribution of the node) comprise the node's Markov blanket.

CSI is another important inference property in BNs, and less restrictive than CI. The idea behind it is that certain independence relations may happen under certain realizations of RVs, i.e. only when certain RV values are observed. In such scenarios, even if CI is not present the \emph{context} of the RVs would allow to perform inference. This less restrictive context 
arises more frequently than CI, particularly in relational models \cite{Fierens:10}. Below, we adapted the definition of CSI from \cite{Boutilier:96} and \cite{Fierens:10}. 



\begin{definition}\textbf{Context-specific independence:}
Let $\mathbf X$, $\mathbf Y$ and $\mathbf W$ be distinct sets of RVs. Then $\mathbf X~{\independent}_c~\mathbf Y~|~\mathbf W=\mathbf w$ (which reads as follows: $\mathbf X$ is context-specific independent of $\mathbf Y$ given $\mathbf W=\mathbf w$) if $P(\mathbf X|\mathbf Y,\mathbf W=\mathbf w)=P(\mathbf X|\mathbf W=\mathbf w)$ whenever $P(\mathbf Y,\mathbf W=\mathbf w)>0$. 
\end{definition}

While CI and CSI are properties consistently used for inference in the BN research community, our task is not to infer unobserved RVs. Instead we would like to take advantage of inference mechanisms for realization of RVs. i.e. for sampling. 


\subsection{Generative Network Models}

The goal of GNMs is to generate random networks $G$ from certain network-distribution $P(G)$. One of the most popular mechanisms used to generate $G$ is to produce a matrix of edge-probabilities $\mathcal P$ from which sampling of a network's adjacency matrix is done. Figure \ref{fig:gnm} shows a matrix of edge-probabilities $\mathcal P$ (left) from which a random adjacency matrix is sampled (center), with its corresponding sampled network (right). For example, $\mathcal P[7,8]=P(E_{78})=\pi_{78}$ has a high probability (dark cell, left plot), and the edge $e_{78}$ is sampled (black cell, center plot). Next, we describe two GNMs that are complex enough to incorporate several levels of RVs.

\vspace{2mm}
\noindent{\bf Block two-level Erd\H{o}s-R\'enyi (BTER) model:}
Block two-level Erd\H{o}s-R\'enyi (BTER) model \cite{Seshadhri:12} is a GNM where networks are sampled in three steps. First, a preprocessing step groups nodes of (almost) the same degree in blocks. Second, the so called \emph{phase-1} of the algorithm creates conventional Erd\H{o}s-R\'enyi graphs for each block, i.e. each edge is created independently with equal probability in the block. 
 The number of edges sampled depends on a parameter provided to the algorithm and on the lowest degree node in the block. Last, the blocks are linked using a Chung-Lu  model~\cite{Chung:02}, which is a type of weighted Erd\H{o}s-R\'enyi model.

\vspace{2mm}
\noindent{\bf mixed Kronecker Product Graph Model (mKPGM):}
mKPGM is a generalization of the Kronecker Product Graph Model (KPGM) \cite{leskovec:10}. KPGM generates a matrix of edge-probabilities $\mathcal P$ by $K-1$  Kronecker product of a matrix of parameters $\Theta$, of size $b\times b$, with itself. 
The value of $K$ is such that will lead to the desired target number of nodes, given that $dim(\Theta)=b\times b$ then $b^K=|\mathbf V|$. Once $\mathcal P$ is calculated, the final network is sampled. On the other hand, mKPGM uses parameter tying to capture the characteristics of a network population~\cite{Moreno:10} as will be described in the next paragraph.


\begin{figure}[t]
\centering
\raisebox{57pt}{\begin{tabular}{l}
\raisebox{25pt}{$k=1$}\\
\raisebox{0pt}{$k=2$}\\
\raisebox{20pt}{$\lambda=0$}\\
\raisebox{0pt}{$k=3$}\\
\raisebox{0pt}{$\lambda=1$}
\end{tabular}}
\hspace{-4mm}
\includegraphics[width=1.5in]{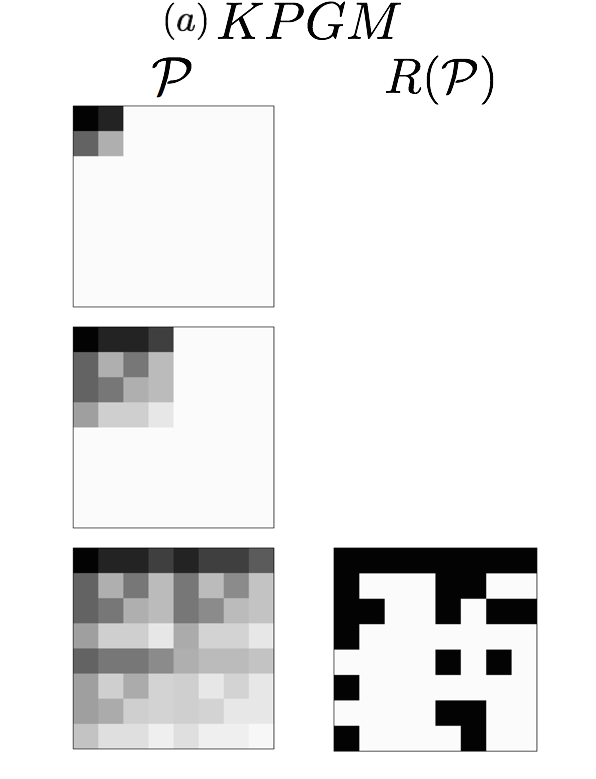}
\hspace{-5.5mm}
\includegraphics[width=1.5in]{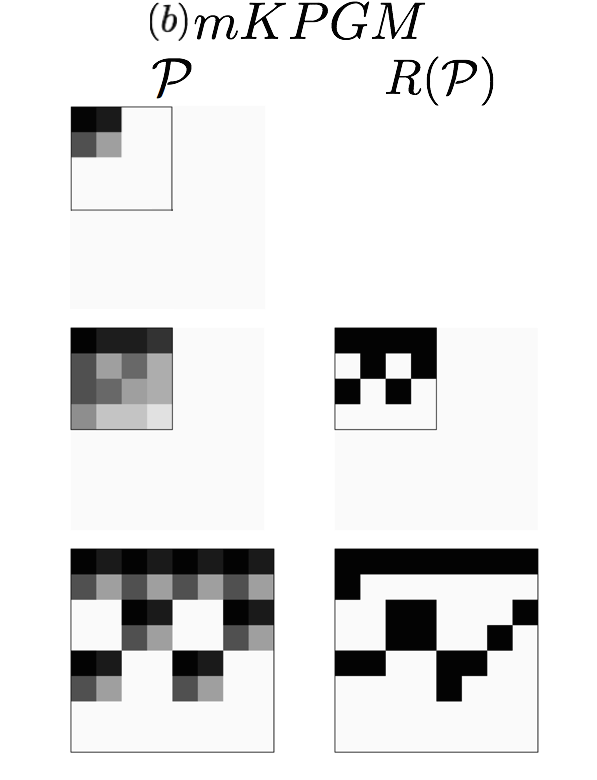}
\caption{KPGM (a) and mKPGM (b) for $K\!=\!3$ and $\ell\!=\!2$.}
\label{figure:egKronModels}
\vspace{-3.5mm}
\end{figure}

\subsection{Sampling from GMNs}

 {\bf mKPGM sampling:}
Given the parameter-matrix $\Theta$  dim$(\Theta)\!=\!b\!\times\! b$ ($\forall\ \! i,\!j \; \theta_{ij}\!\in \![0,1])$, the number of Kronecker multiplications $K$, and the number of untied levels $\ell$, mKPGM  generates a network as follows: First, it computes $\mathcal P^{\ell}$ by $\ell-1$ Kronecker product of $\Theta$ with itself.
Second, it samples a network $G^{\ell}=(\mathbf V^{\ell},\mathbf E^{\ell})$ from $\mathcal P^{\ell}$ by sampling each cell independently from a $Bernoulli(\mathcal P^{\ell}_{ij})$. 
Third, the algorithm calculates $\mathcal P^{\ell + \lambda}\!=\!G^{\ell+\lambda-1}\otimes \Theta$ and samples $G^{\ell+\lambda}$ for $\lambda\!=\!1\ldots K\!-\!\ell$ as before. 
This iterative process, of Kronecker multiplications and sampling, ties parameters and increases the variability over the generated network of the model. 
$\lambda$ references a tying iteration in the mKPGM sampling process.
We will refer to each cell sampled with mKPGM as an RV with Bernoulli distribution. Notice that this RVs represent edges in the last tying iteration of the mKPGM sampling process and sets of edges (clusters) at higher levels of the mKPGM tying iterations. 

Figure~\ref{figure:egKronModels} shows an example of KPGM and mKPGM with parameters $K\!=\!3$, $\ell\!=\!2$, $b=2$, and $\Theta=\left[\begin{tabular}{cc}0.9 & 0.7 \\ 0.5 & 0.3\end{tabular}\right]$. KPGM generates the probability matrix $\mathcal P$ (left column $k\!=\!3$) before sampling the final network (right column $k=3$). Instead, mKPGM sample $G^{\ell}$ at $k\!=\!2\!=\!\ell$. Then, it generates $\mathcal P^{3}\!=\!G^{2}\otimes \Theta$ and samples $G^{3}$ for $\lambda\!=\!1$.

\vspace{2mm}
\noindent\textbf{Group Sampling:}
Group Probability sampling (GP) is a general sampling method that can be applied to many types of GNMs. It is an alternative to the normal sampling approach of most GNM where edges are sampled one-by-one. Instead, GP allows to sample groups of edges all sharing the same probability of being sampled.
GP is an unbiased, provably correct, and efficient sampling process that can be applied to any GNMs that define a matrix $\mathcal P$ of edge-probabilities. 
Given a GNM with parameter $\Theta$ that defines $\mathcal P$, GP samples a network in three steps. 
First, 
it derives $\mathbf U$ a set of unique probabilities ($\pi_k$) in $\mathcal P$ as determined by the GNM. 
Second, for each $\pi_k\in\mathbf U$ it calculates $T_k$, the number of possible edges associated with $\pi_k$, and samples the number of edges $x_k$, to be placed among $T_k$ possible ones with $P(X_k = x_k)\!\sim\!Bin(n,p)\Rightarrow\!n\!=\!T_k,p\!=\!\pi_k$ (because the number of successes in $T_k$ Bernoulli trials with probability $\pi_k$ are binomial-distributed).
Third, it samples $x_k$ edges at random among the $T_k$ possible edges with probability $\pi_k$. 
This process can be applied to each tied iteration $\lambda$ of the mKPGM model.
For further details of the GP sampling for mKPGM, please refer to~\cite{Moreno:14}.


\begin{figure}[]
\includegraphics[width=3.25in]{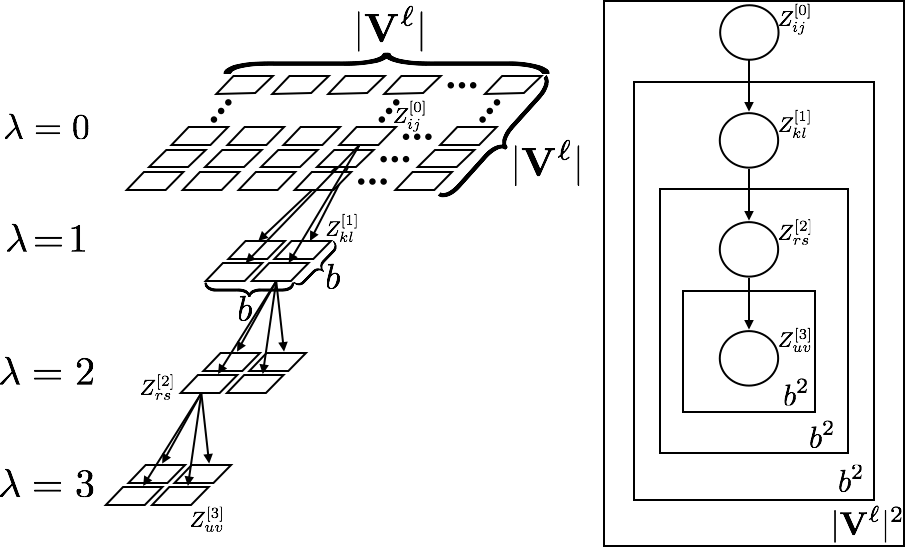}
\caption{Left: RVs of an mKPGM sampling process. Right: plate notation BN equivalence of the same mKPGM RVs.}
\label{fig:BN-mKPGM}
\vspace{-4mm}
\end{figure}

\section{Generative Network Models \\Represented as Bayesian Networks}

Bayesian networks can be used to represent the relationships between RVs in GNMs. As we mentioned in the introduction, for some GNMs, since the edge RVs are independent, it is unnecessary to consider a BN representation. For example, the model in Figure~\ref{fig:gnm} corresponds to an 8-node undirected network with no-self loops, thus there are $28$ independent edge RVs. 
However, a BN representation is more appropriate for new models with more complex dependencies among the edges (such as mKPGM and BTER), and inference or sampling can be done based on the associated graphical models. 


\begin{remark}
\label{thm:BNmKPGM}
An mKPGM model $\mc M$ with parameters $\Theta$, $K$, and $\ell$ can be represented as a BN $\mc N$ with a tree structure and parameters $\Theta'$ obtained from $\Theta$: $\mc M_{\Theta}{} \reducesto \mc N_{\Theta'}$. 
\end{remark}

%

The mKPGM model consists of multiple levels of RVs. 
The first of these levels corresponds to the Bernoulli parameters in $\mc P^{\ell}$, the probability matrix to generate the subnetwork $G^{\ell}\!=\!(\mathbf V^{\ell},\mathbf E^{\ell})$. Each possible $E^{\ell}_{ij} \in \mathbf E^{\ell}$ is generated with probability $P(E^{\ell}_{ij})=\mc P^{\ell}[i,j]$. With a $b \times b$ parameter matrix $\Theta$, there are $(b^\ell)^2=|\mathbf V^{\ell}|^2=|\mc P^{\ell}|$ possible edges. 

These potential edges, at the top of the hierarchy, can be modeled as independent RVs in a BN (i.e., the root nodes of the BN).
Let $Z^{[0]}_{ij}$ be the RV in the BN representing the edge $E^{\ell}_{ij}$. Then the BN representation of this level of the hierarchy corresponds to $(b^\ell)^2$ independent RVs, with:
\begin{align}
P(Z^{[0]}_{ij}=1)&=\mc P^{\ell}[i,j] \nonumber \\
P(Z^{[0]}_{ij}=0)&=1- \mc P^{\ell}[i,j] \nonumber
\end{align}

More generally, we will use the notation $Z^{[\lambda]}_{ij}$ to refer to RVs in the BN representation, where $\lambda=[0, K\!-\ell]$ refers to the level of tying in the mKPGM. The first level $(\lambda=0$) refers to the untied portion of the mKPGM. For notational purposes, we will use $\mathbf Z^{[0]}$ to refer to the set of all RVs $Z^{[0]}_{ij}$. 

The next level corresponds to the Kronecker product of $G^{\ell}$ with $\Theta$, which produces $\mc P^{\ell+1}\!=\!G^{\ell}\otimes\Theta$. There are $(b^{(\ell+1)})^2=|\mc P^{\ell+1}|$ possible edges in the next level of the hierarchy, with each edge $E^{\ell}_{ij}$ impacting $b^2$ of the edges in $\mathbf E^{\ell + 1}$ due to the Kronecker product (i.e., $E^{\ell + 1}_{kl}$ is generated from $\mc P^{\ell + 1}_{kl} \!=\! E^{\ell}_{ij} \theta_{xy}$ for some $i,j \!\in\! [1,b^\ell]$ and $x,y \!\in\! [1,b]$ s.t. $\theta_{xy} \in \Theta$).

The BN representation of this level of the hierarchy consists of a random variable $Z^{[\lambda=1]}_{kl}$ for each edge $E^{\ell+1}_{kl}$, for a total of $(b^{(\ell+1)})^2$ RVs. The  Kronecker product relationships are modeled by dependencies in the BN, so each $Z^{[0]}_{ij} \in \mathbf Z^{[0]}$ has $b^2$ descendants in $\mathbf Z^{[1]}$. Thus the RVs in $\mathbf Z^{[1]}$ can be thought of as $|\mathbf V^{\ell}|^2$ sets of RVs, each of size $b^2$, which share a common parent in $\mathbf Z^{[0]}$. For an edge $E^{\ell + 1}_{kl}$ that is generated via $E^{\ell}_{ij} \theta_{xy}$, the conditional probability for its associated RV is:
\vspace{-2mm}
\begin{align}
P(Z^{[1]}_{kl}=1 | Z^{[0]}_{ij}=1 )&=  \theta_{xy}\nonumber \\
P(Z^{[1]}_{kl}=0 | Z^{[0]}_{ij}=1 )&=1- \theta_{xy} \nonumber \\
P(Z^{[1]}_{kl}=1 | Z^{[0]}_{ij}=0 )&=  0 \nonumber \\
P(Z^{[1]}_{kl}=0 | Z^{[0]}_{ij}=0 )&= 0 \nonumber
\end{align}

The remaining levels of the mKPGM can be transformed by the same process. 
In general, a level $\lambda$ of the mKPGM hierarchy is represented by a set of $(b^{(\ell+\lambda)})^2$ RVs in $\mathbf Z^{[\lambda]}$, where $b^{\ell+\lambda}$ is the number of nodes in the graph $G^{\ell+\lambda}$. Each $Z^{[\lambda]}_{kl} \!\in\! \mathbf Z^{[\lambda]}$ has one parent in $\mathbf Z^{[\lambda-1]}$ and each $Z^{[\lambda-1]}_{ij} \!\!\in\! \mathbf Z^{[\lambda \!-1]}$  has $b^2$ descendants in $\mathbf Z^{[\lambda]}$.

This process generates a tree structure where groups of $b^2$ RVs have the same parent in $\mathbf Z^{[\lambda-1]}$. Two variables $Z^{[\lambda]}_{ij}$ and $Z^{[\phi]}_{ij}$ at levels $\lambda$ and $\phi$ are dependent if they share a common ancestor. 

The final BN $\mc N $ consists of all the RVs $\mathbf Z^{[0]}, \mathbf Z^{[1]}, ..., \mathbf Z^{[\lambda = K\!-\ell]}$ and their associated probabilities.
This shows that the BN $\mc N $ represents the model $\mc M$, i.e. $\mc M \reducesto \mc N$. 

An example BN representation of an mKPGMs is visualized in Figure~\ref{fig:BN-mKPGM} for $\lambda={0,1,2,3}$. Here $\lambda=0$ corresponds to $G^{\ell}$ in the mKPGM sampling process. 
There is a total of $(b^{\ell})^2\!=\!|\mathbf V^{\ell}|^2$ RVs each of them represented by a $Z^{[0]}_{ij}$. Note the use of double subindex for the $Z$ RVs is to indicate the position of the RV in the cluster/edge matrix. Each of these RVs has $b^2$ descendants at $\lambda=1$. However, to make it easier visualize the relations among the variables in the left subplot, we drop the descendants for all RVs except one in each level of the hierarchy. In the right subplot, the descendants are represented more generally by the plate notation. 


We note that the tree structure of the GNM-associated BN, along with the recursive nature of the GNM and the symmetries among RVs with the same probability, would make it amenable for lifted-inference. However, for this paper our discussion is centered in the problem of sampling.

\section{Sampling from Bayesian Networks}
Given that an mKPGM can be reduced to a BN, we now consider sampling from the associated BN to generate a network from the underlying mKPGM model. The process to sample from an empty BN is straightforward. It involves determining a topological sorting of the RVs, then iteratively sampling a value for each RV conditioned on the sampled values of its parents. We will discuss how the structure of the associated BN can be exploited to speed up this process below. However, we note that the complexity increases if the sampling is conditioned on evidence and the BN representation will facilitate even further gains for these more complex inference tasks.




\subsection{Naive Sampling Using Conditional Independence }
Given that the BN for mKPGMs is tree-structured, it is easy to determine a topological sort that will facilitate sampling. Specifically, each tree rooted at an RV in $\mathbf Z^{[0]}$ is independent of the others. Moreover, within a particular tree, at level $\lambda$, each $Z^{[\lambda]}_{ij}$ is conditionally independent of the others ($\mathbf Z^{[\lambda]} - \{Z^{[\lambda]}_{ij}\})$ given the value of its parent in $\mathbf Z^{[\lambda-1]}$.
Thus it is simple to use the hierarchy itself as the topological ordering for sampling. Given that the RVs at level $\lambda$ of the hierarchy are conditionally independent once all the RVs from $\lambda-1$ are sampled, the order in which the RVs are sampled within the same level is not important. Furthermore, since each CPT corresponds to a $2 \times 2$ matrix (where if the parent value is zero the RV has zero probability of being sampled, otherwise it has a probability equal to some $\theta_{xy}\in\Theta$), sampling of each RV value is constant.  Thus, the complexity of sampling will be a function of the number of RVs in the BN. 
Unfortunately the number of RVs increase at each level of the hierarchy. The number of RVs at hierarchy $\lambda$ is equal to $(b^{\ell+\lambda})^2$ so this results in a total number of RVs:
\[ \sum_{\lambda=0}^{K-\ell}(b^{\ell+\lambda})^2=(b^2)^\ell\sum_{\lambda=0}^{K-\ell}(b^2)^{\lambda}=\frac{(b^2)^{K+1}-(b^2)^\ell}{b^2-1} \] 

\noindent which is significantly larger than the number of possible edges in the network: $N^2 = b^K$.

\subsection{Context-Specific Independence for Network Sampling}
Context-specific independence (CSI) could be used to improve sampling efficiency by either reducing the size of the CPTs or simplifying the ordering of RVs (e.g., facilitating parallelization). 

To exploit CSI, we first need to identify the context in the mKPGMs for which independence between random variables arises. 
Recall that for three RVs $X, Y, Z$, the definition of CSI is $X~{\independent}_c~ Y~|~ W\!=\!w$ if $P( X| Y, W\!=\! w)=P( X| W\!=\! w)$. Since each RV in the mKPGM BN  has a single parent, with the topological ordering discussed above there is not any opportunity to use CSI to improve the efficiency of the sampling process. However, CSI could be useful for more complicated inference tasks that condition on evidence.
\section{Context-Specific Dependence \\for Network Sampling}


We now formalize the concept of context-specific dependence (CSD). Note that in the definition, $\mathbf W$ can be any set of RVs in a BN and is not necessarily related to the RVs for mKPGM. 

\begin{definition}\textbf{Context-specific dependence:}
Let $\mathbf X$, $\mathbf Y$ and $\mathbf W$ be distinct sets of RVs. Then $\mathbf X~{\notindependent}_c~\mathbf Y~|~\mathbf W\!=\!\mathbf w$ if $P(\mathbf X|\mathbf Y,\mathbf W\!=\!\mathbf w)\neq P(\mathbf X|\mathbf W\!=\!\mathbf w)$ whenever $P(\mathbf Y,\mathbf W\!=\!\mathbf w)>0$. 
\end{definition}

Both CSI and CSD may appear in GNMs graphical models. Whenever independence of RVs in a BN appear due to specific context, then CSI properties can be exploited to {\em relax} the constraints on inference and sampling. On the other hand, the BN representation itself generally implies CSD---since it is assumed that an RV depends on the value of its parents. However, if the CSD produces more structure (e.g., additional symmetry, more extreme dependence) then its properties can be exploited to {\em tighten} the constraints on inference and sampling.

In GNMs, the BN structure has a more specific dependency that can be used for efficient sampling:


\begin{definition}\textbf{Deterministic CSD (DCSD) in mKPGMs:} 
Let $\mathcal{M}$ be an mKPGM with associated BN $\mathcal{N}$. Let $P(Z_{ij}^{[\lambda]})$ be the probability in $\mathcal{N}$ that the RV $Z_{ij}^{[\lambda]}=1$.
$\mathcal N$ is {\em deterministic context-specific dependent}  if at each layer $\lambda$, it partitions all RVs $Z_{ij}^{[\lambda]}$, such that:
$$
 P\left(Z_{ij}^{[\lambda]}=1\left|pa(Z_{ij}^{[\lambda]})=0\right.\right)=0~~~\forall~i,j,\lambda
 $$
\noindent where
$P\left(Z_{ij}^{[\lambda]}\!=\!1 \left|pa(Z_{ij}^{[\lambda]})\!=\!1\right.\right)> 0~\forall~i,j,\lambda$.
\end{definition}

Combining the hierarchical order sampling process discussed previously and DCSD, we can reduce the complexity of sampling a network. Specifically, once the $|\mathbf V^{\ell}|^2$ RVs are sampled from the first hierarchy level ($\lambda=0$), instead of sampling all variables of the second level ($\mathbf Z^{[1]}$), we avoid considering the RVs with parent values of zero. This results in a considerable reduction in the number of sampled RVs, which is propagated down the hierarchy. For example, if $Z^{[0]}_{ij}\!=\!0$, we avoid sampling $(b^2)^{K-\ell}$ RVs (i.e., $b^2$ descendants are recursively affected at each of the $\lambda=K\!-\!\ell$ levels). Let $N_Z^{[\lambda]}$ be the number of active RVs (i.e., value of 1) at layer $\lambda$. Then the number of variables to be sampled in the next level is equal to $N_Z^{[\lambda]} \cdot b^2$ (each variable has $b^2$ descendants). As demonstrated in previous work on mKPGMs, the expected number of edges at layer $\lambda$ is $N_Z^{[\lambda]} \!=\! \left(\sum\Theta\right)^{\ell+\lambda}$ \cite{Moreno:10}.  
Thus, in expectation, the total number of RVs sampled using DCSD is $\displaystyle \sum_{\lambda=0}^{K-\ell}N_Z^{[\lambda]}$. Also, since the RVs we only analyze random variables with active parent, the CPT look up can be reduced to a single value. These simplifications produce a considerable reduction in the time complexity of the network sampling process.


It is important to note that exploiting DCSD for mKPGM sampling will generate networks from the true network distribution as long as GP sampling is applied to randomly sample from RVs with the same probability at each tied iteration. This is because GP sampling generates networks from the true network distribution \cite{Moreno:14}. 
\section{Complexity Analysis Comparison}
As stated before, the sampling process is the same for all BN regardless of the method used: CI or DCSD. This process involves determining a topological sorting of the RVs, then iteratively sampling a value for each RV conditioned on the sampled values of its parents. Consequently, the difference in performance between the different methods depends on two factors: the number of RVs to be sampled, and the complexity of the CPT look up to sample from the RVs.

Table \ref{table:num-rvs} shows a comparison of the number of sampled RVs and the number of parent combinations in the CPTs for the sampling methods discussed in the paper. Recall that $b$ corresponds to the size of the original parameter matrix $(dim(\Theta)=b\times b)$, $K$ defines the number of Kronecker products, $\ell$ is the number of independent hierarchy levels for mKPGM, and thus $\lambda\in\{0,\dots,K-\ell\}$.

DCSD allows more efficient sampling than CI because the number of RVs is smaller than CI:
$\frac{(b^2)^{K+1}-(b^2)^\ell}{b^2-1} > \sum_{\lambda=0}^{K-\ell}N_Z^{[\lambda]}$. This is easy to verify.
Assuming each entry of $\Theta$ with size $b\times b$ is a valid probability and hence $\Theta_{ij}<1$, then $b^2 > \sum \Theta$. Then, $\sum_{\lambda=0}^{K-\ell}(b^2)^{\ell+\lambda} > \sum_{\lambda=0}^{K-\ell}\left(\sum\Theta\right)^{\ell+\lambda}$.

It is worth noticing the relation of the number of possible edges $N_v^2=b^K$ and the number of RVs in CI and DCSD. $N_v^2$ is equal to the last term of $\sum_{\lambda=0}^{K-\ell}(b^2)^{\ell+\lambda}$. 
On the other hand, the last term of $\sum_{\lambda=0}^{K-\ell}N_Z^{[\lambda]}$ is $(\sum \Theta)^{K}<N_v^2$.

Finally, most real networks are sparse, which means $|\mathbf E| = O(N_v)=b^K$. However, the number of RVs using CI is larger than $N_v^2$. In expectation, each level of the mKPGM hierarchy will sample $O(b^{\ell+ \lambda})$ edges. The total number of sampled RVs is bounded by $ \sum_{\lambda=0}^{K-\ell}N_Z^{[\lambda]} \cdot b^2<b^{K+2} \sum_{\lambda=0}^{K-\ell} 1<(K-\ell+1)b^{K+2}$. This bound in expectation (ebound) is significantly less than $N_v^2$ 



\begin{table}
\centering
\begin{tabular}{|l|l|l|}
\hline
Property &Number of RVs & $pa$ values\\
\hline
CI &$\displaystyle\frac{(b^2)^{K+1}-(b^2)^\ell}{b^2-1}$ & 2\\
\hline
DCSD&$\displaystyle\sum_{\lambda=0}^{K-\ell}N_Z^{[\lambda]} $ & 1\\ \hline
ebound DCSD&$\displaystyle(K-\ell+1)b^{K+2}$ & 1\\
\hline
\end{tabular}
\caption{Complexity for GMNs sampling methods that exploit different properties of the associated BN.}\label{table:num-rvs}
\vspace{-3.9mm}
\end{table}

\section{Discussion, Current and Future Work}

CSI and CSD are complementary properties arising in graphical models, in which the context changes the constraints during inference---either by relaxing or tightening the constraints. By identifying and taking advantage of these properties, it is possible to perform more efficient inference and sampling. 

We showed an example of a GNM that can be reduced to a graphical model and that sampling could be done from multiple perspectives. While sampling efficiencies based on CSI are not available for this type of BN, exploiting DCSD allows us to develop a faster sampling process (compared to conventional CI sampling). This improvement is primarily due to a reduction in the number of sampled RVs. Combined with group sampling, DCSD properties can be exploited for fast and provably correct sampling in other GNMs with complex dependencies, as in mKPGM.
However, in mKPGMs the DCSD properties may also complicate inference tasks that condition on evidence---because the nature of DCSD constrains the problem and reduces the number of possible solutions. The implications of this are the subject of our ongoing work.


\section{ Acknowledgments}
This research is supported by NSF and DARPA under contract numbers IIS-1149789, CCF-0939370, and N660001-1-2-4014. The U.S. Government is authorized to reproduce and distribute reprints for governmental purposes notwithstanding any copyright notation hereon. 

\bibliography{neville-all} 
\bibliographystyle{aaai}

\end{document}